\documentclass[letterpaper]{article} 
\usepackage{aaai2026}  
\usepackage{times}  
\usepackage{helvet}  
\usepackage{courier}  
\usepackage[hyphens]{url}  
\usepackage{graphicx} 
\usepackage{natbib}  
\usepackage{caption} 
\usepackage{algorithm}
\usepackage{algorithmic}

\usepackage{booktabs}
\usepackage{multirow}
\usepackage{array}
\usepackage{pifont}            
\newcommand{\cmark}{\ding{51}} 
\newcommand{\xmark}{\ding{55}} 
\usepackage{amsmath} 
\usepackage{upgreek}

\usepackage{newfloat}
\usepackage{listings}
\DeclareCaptionStyle{ruled}{labelfont=normalfont,labelsep=colon,strut=off} 
\floatstyle{ruled}
\newfloat{listing}{tb}{lst}{}
\floatname{listing}{Listing}
\title{PEOD: A Pixel-Aligned Event-RGB Benchmark for Object Detection \\ under Challenging Conditions}
\author{
   Luoping Cui\equalcontrib\textsuperscript{\rm 1},
   Hanqing Liu\equalcontrib\textsuperscript{\rm 1},
   Mingjie Liu\equalcontrib\textsuperscript{\rm 1},
   Endian Lin\textsuperscript{\rm 1},
   Donghong Jiang\textsuperscript{\rm 1},
   Yuhao Wang\textsuperscript{\rm 1},
   Chuang Zhu\textsuperscript{\rm 1}\thanks{Corresponding author.}
}
\affiliations{
    \textsuperscript{\rm 1}School of Artificial Intelligence, Beijing University of Posts and Telecommunications, Beijing, China\\
  \{ lpcui, hanqingliu, LMJ, ledgogo, donghongjiang, wyhao, czhu \}@bupt.edu.cn

}

\usepackage{bibentry}

\begin{document}

\maketitle

\begin{abstract}

Robust object detection for challenging scenarios increasingly relies on event cameras, yet existing Event-RGB datasets remain constrained by sparse coverage of extreme conditions and low spatial resolution ($\leq 640 \times 480$ ), which prevents comprehensive evaluation of detectors under challenging scenarios. To address these limitations, we propose PEOD, the first large-scale, pixel-aligned and high-resolution ($1280 \times 720$) Event-RGB dataset for object detection under challenge conditions. PEOD contains 130+ spatiotemporal-aligned sequences and 340k manual bounding boxes, with 57\% of data captured under low-light, overexposure, and high-speed motion. Furthermore, we benchmark 14 methods across three input configurations (Event-based, RGB-based, and Event-RGB fusion) on PEOD. On the full test set and normal subset, fusion-based models achieve the excellent performance. However, in illumination challenge subset, the top event-based model outperforms all fusion models, while fusion models still outperform their RGB-based counterparts, indicating limits of existing fusion methods when the frame modality is severely degraded. PEOD establishes a realistic, high-quality benchmark for multimodal perception and facilitates future research.

\end{abstract}

\begin{links}
    \link{Datasets}{https://github.com/bupt-ai-cz/PEOD}
\end{links}

\section{Introduction}
\begin{figure}[t]
    \centering
    \includegraphics[width=0.45\textwidth]{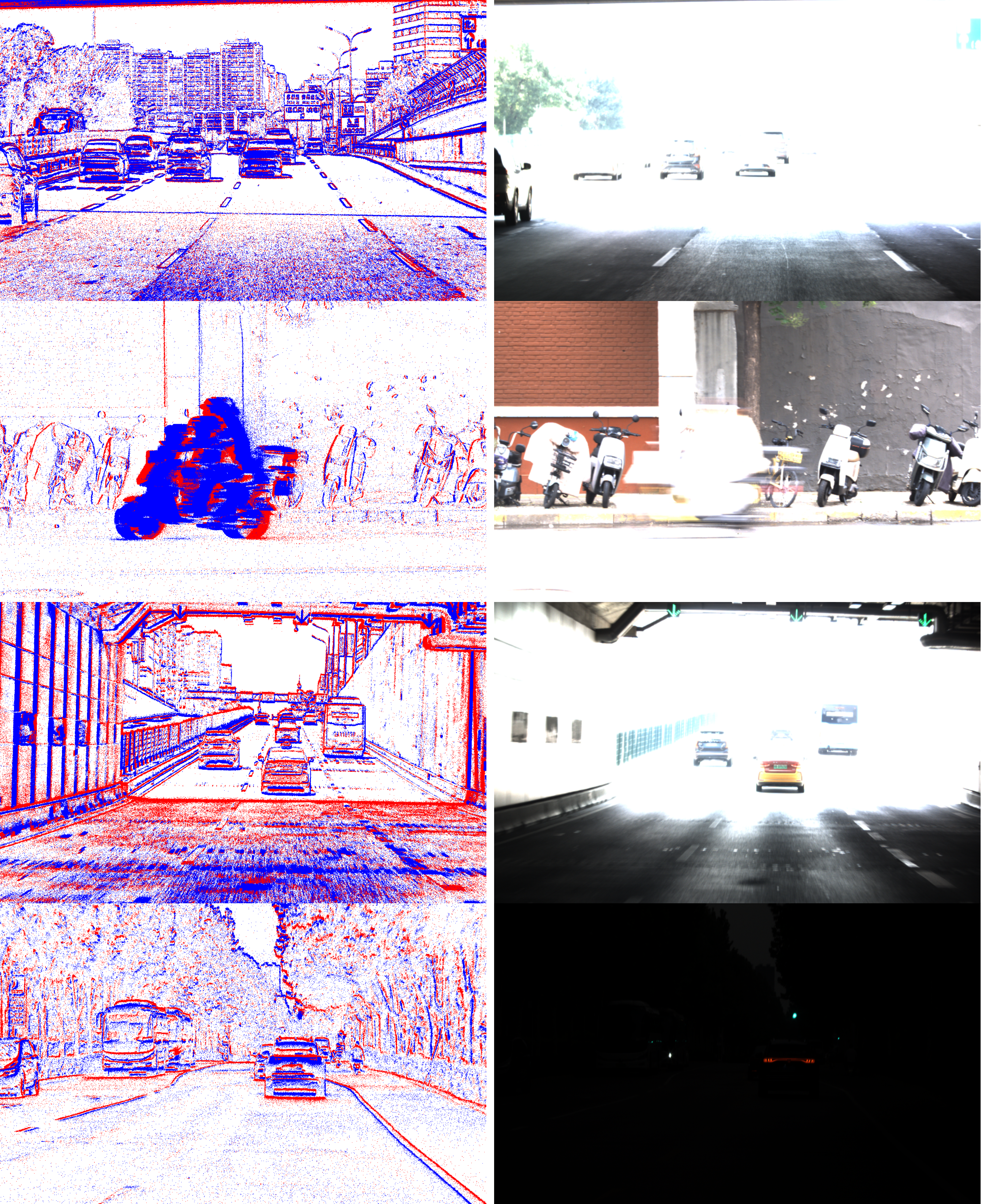}
    \caption{\textbf{PEOD examples under diverse challenging conditions.} Overexposure (rows 1–3), motion blur (row 2), and low-light (row 4). Each row presents the event stream (left) with its pixel-aligned RGB frame (right).}
    \label{fig:datasetshow}
\end{figure}

Object detection is a critical perception task for intelligent systems, including robotics \cite{introFalanga2020}, autonomous vehicles \cite{Teng2023intr}, and surveillance \cite{Wei2021intrT,Du2023intT}. However, conventional frame-based cameras suffer from inherent limitations in exposure time and dynamic range, leading to low-quality images and significant information loss in challenging scenarios such as high-speed motion and varying illumination \cite{Frame_noDai2023}. 
Event cameras, which asynchronously report per-pixel brightness changes, offer a new paradigm with their microsecond-level temporal resolution and high dynamic range, demonstrating superior performance in extreme conditions \cite{Prophesee2024}. However, event data lack the rich texture and static scene information that are a core strength of RGB cameras. Consequently, fusing RGB frames and event streams is a highly promising approach for building robust, all-day, all-scenario detection systems \cite{DSEC_DET2024}.

While existing dual-modality datasets such as DSEC \cite{dsecGehrig2021}, PKU-SOD \cite{Li2023}, have enabled initial research, they suffer from critical limitations: \textbf{1) \textit{Scarcity of Extreme Scenarios} }: Challenging data (e.g., night, overexposure, motion blur) constitutes less than 20\% of existing datasets, hindering the proper evaluation of model robustness. \textbf{2) \textit{Low Resolution}}: Most datasets feature resolutions like $346 \times 260$ or $640 \times 480$, which are insufficient for modern detectors that require fine-grained detail. 

To overcome these bottlenecks, we introduce a new, large-scale pixel-aligned Event-RGB dataset for object detection (PEOD), as Figure \ref{fig:datasetshow}. It is the first dataset of its kind, captured with a high-resolution ($1280 \times 720$) EVK4 event camera \cite{Prophesee2024} and an RGB camera, using a beam-splitter optical system and a hardware signal generator to achieve spatialtemporal synchronization. The dataset covers diverse driving environments, with over 57\% of the data captured in extreme conditions, including low-light, overexposure, and high-speed motion.We establish a comprehensive benchmark to unify the evaluation of existing event and Event-RGB fusion algorithms on PEOD dataset. Our main contributions can be summarized as follows:

\begin{itemize}
\item We introduce a first-of-its-kind dataset containing more than 57\% of the data collected under extreme conditions (low-light, overexposure, and high-speed motion).
\item We provide the first high-resolution ($1280 \times 720$), pixel-aligned Event-RGB dataset, with 340k manually annotated bounding boxes in six traffic-related object classes.
\item We establish a comprehensive benchmark by evaluating 14 classical and state-of-the-art object detectors under RGB-based, event-based, and Event-RGB fusion settings on both the full dataset and challenging subsets.

\end{itemize}

\section{Related Work}
\subsection{Event and Event-RGB Vision Datasets}
Numerous event-based datasets have been introduced to address a diverse set of vision tasks, including object detection, object tracking, and anomaly detection. For object detection task, Gen1 dataset \cite{Gen12020} offers limited resolution, but only annotates two object categories. The 1 Mpx dataset offers event streams accompanied by annotations mapped from frames. eTram dataset \cite{Verma2024} focuses on monitoring of traffic scenes, providing an all-day dataset. PEDRo dataset \cite{Boretti2023} is dedicated to pedestrian detection in service robotics. DSEC dataset \cite{dsecGehrig2021} employs synchronized Gen3.1 stereo event cameras along with RGB, LiDAR, and RTK-GPS, collecting data across diverse environments. PKU-SOD dataset \cite{Li2023}, collected with a DAVIS346 sensor, is a large-scale Event–RGB benchmark with three-class annotations. For object tracking task, EventVOT dataset \cite{Wang2024} provides a large-scale, high-resolution benchmark, comprising 1,141 videos that cover 19 object categories. For 3D perception tasks, MVSEC dataset \cite{MVSECZhu2018} extends to various platforms, and provides synchronized IMU and LiDAR ground truth. For gesture recognition task, EvRealHands dataset provides a real-world event-based resource for 3D hand pose estimation. \cite{EvHandPoseJiang2024}. For anomaly detection task, UCF-Crime-DVS \cite{Qian2025} dataset captures event streams for video anomaly detection based on the UCF-Crime dataset. For image reconstruction task, RLED dataset \cite{Liu2024recon} uses a coaxial imaging setup to capture 64k synchronized RGB frames and event streams, offering a benchmark for nighttime reconstruction. 

Existing event and Event-RGB datasets, although valuable, are still constrained by low spatial resolution, sparse annotations, or a narrow sampling of scenes captured under challenging illumination, which together limit their utility for comprehensive benchmarking. Therefore, our goal is to construct a dataset that simultaneously provides high spatial resolution, densely annotated ground truth, and large-scale coverage of adverse lighting scenarios.

\subsection{Object Detection with Event Cameras}
Research in event-based object detection now follows two main directions: event-based detectors and Event-RGB fusion detectors. Event-based detectors range from CNN slice-based approaches \cite{Li2022,Fan2024CNN} to Transformer variants such as RVT \cite{Gehrig2023} and SAST \cite{Peng2024}, as well as SNN-based models \cite{Luo2024,Fan2024}, yet all remain constrained by an inherent insensitivity to texture details and a limited ability to detect slow-moving or stationary objects in event data. These inherent limitations motivate the second paradigm: Event-RGB fusion. Fusion detectors overcome these limitations by combining dense RGB texture information with event-derived motion cues \cite{DSEC_DET2024}. Initial fusion methods rely on a cascade policy of converting event streams into pseudo-frames for simple concatenation with RGB features, which offers marginal improvements in robustness \cite{Tomy2022}. To facilitate more effective feature interaction, attention mechanisms have been introduced into fusion frameworks, employing temporal Transformers and asynchronous cross-modal attention for bimodal feature integration \cite{Li2023}. Moreover, the incorporation of advanced strategies like multi-scale feature aggregation, bi-directional calibration, and illumination-adaptive compensation has further boosted model accuracy and robustness \cite{Liu2024}.

Overall, while event-based detectors excel in handling challenging scenarios such as extreme lighting and motion blur, they still face limitations in detecting texture details and static objects. By exploiting complementary information from RGB frames and event streams, Event-RGB fusion detectors achieve robust detection performance across diverse illumination  conditions.

\begin{figure*}[htbp]
    \centering
    \includegraphics[width=1\textwidth]{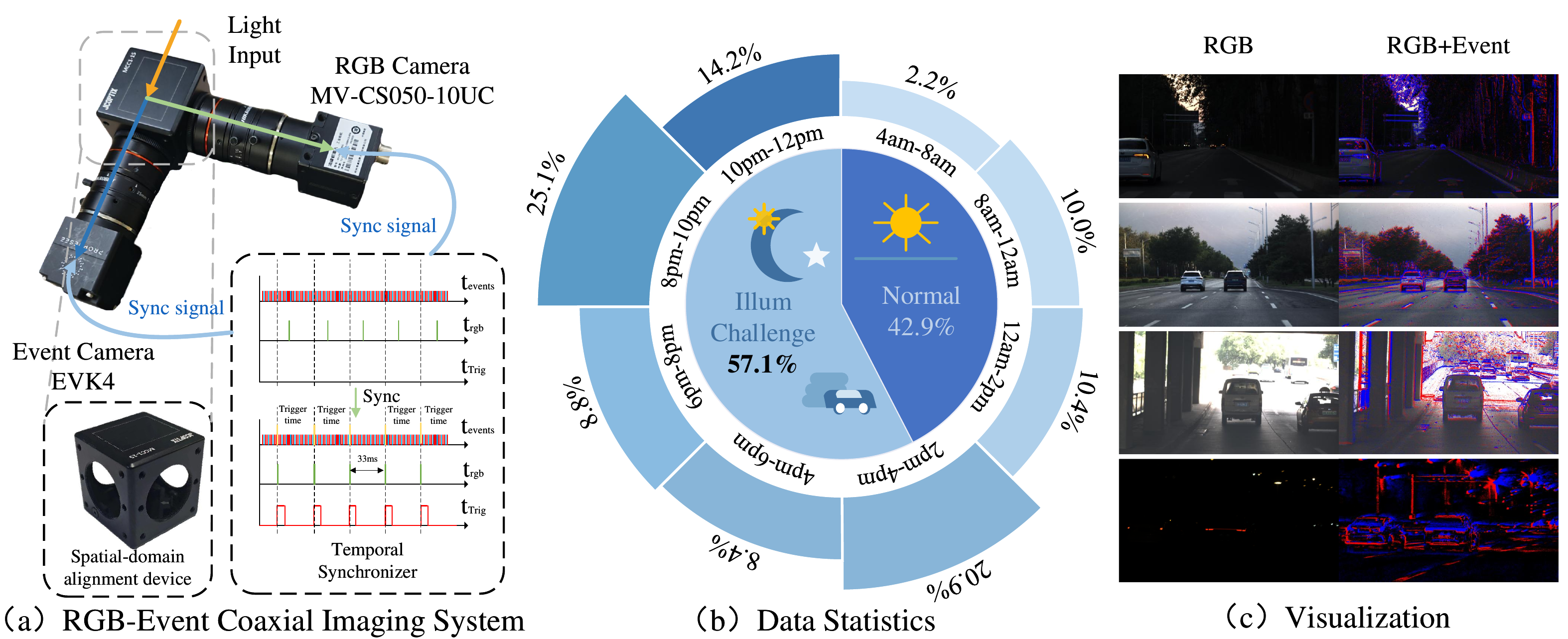}
    \caption{ \textbf{Overview of PEOD dataset and acquisition system.} (a) The coaxial imaging system used to capture spatiotemporally aligned Event and RGB data. (b) Temporal distribution of the dataset, with 57.1\% captured under challenging illumination conditions. (c) Sample aligned Event-RGB pairs from diverse driving scenarios. }
    \label{fig:datasta}
\end{figure*}

\section{PEOD Dataset}
In this section, we systematically introduce a high-resolution, \textbf{P}ixel-aligned \textbf{E}vent-RGB \textbf{O}bject \textbf{D}etction Dataset under challenging scenarios. We provide a comprehensive analysis of our data acquisition system, collection methodology, and the composition of dataset.

\subsection{Dual-Camera Sync System}
To construct a high-resolution Event-RGB dataset with strict spatiotemporal alignment and a unified imaging scale, we developed a coaxial optical system. This system comprises a JCOPTIX OSB25R55-T5 non-polarizing plate beam splitter (50:50 split ratio) and an MCC1-1S $10mm$ coaxial cube, as illustrated in Figure \ref{fig:datasta} (a). This configuration allows an event camera and an RGB camera to share the same optical path, enabling pixel-level spatial alignment through a standard stereo rectification procedure. For precise temporal synchronization, a single square-wave signal generator provides hardware trigger pulses to both cameras, achieving microsecond-level accuracy. The event stream is captured by a Prophesee EVK4 HD camera ($1280 \times 720$), while RGB frames are synchronously acquired by a Hikvision MV-CS050-10UC industrial camera ($2448 \times 2048, 60FPS$). To eliminate discrepancies in focal length and lens distortion, we equipped both cameras with identical Hikvision $25mm$ C-mount fixed-focal-length lenses and maintained a fixed aperture setting for all recordings.

\subsection{Data Collection and Annotation}
Using the acquisition system mounted behind a car's front windshield, we recorded all sequences at 30Hz. Data were collected continuously from 04:00 h to 24:00 h, covering lighting conditions that range from dawn to nighttime, and across diverse environments such as urban roads, suburban roads, complex intersections, tunnels, and highways.

Given the prevalence of challenging conditions such as high speeds, low-light, and overexposure, we adopted a hybrid annotation strategy to ensure the accuracy of the label. For normal conditions, annotations were performed directly on the high-quality RGB frames. For challenging conditions, we leveraged an advanced reconstruction algorithm, NER-Net, to generate grayscale images from the asynchronous event streams, matching the RGB camera's frequency. Annotations were then carried out on these clear, reconstructed frames. Under normal lighting, bounding boxes were annotated directly on the raw RGB frames. Under challenging conditions, we directly annotated the high-clarity reconstructions generated by NER-Net\cite{Liu2024recon}. Our team manually annotated six common classes (car, bus, truck, two-wheeler, three-wheeler, and person), each of which appears in more than 30\% of the recorded instances. All annotations underwent a rigorous cross-checking review process to ensure high quality and consistency.

{
\setlength{\tabcolsep}{4pt}
\begin{table*}[t]
\centering
\begin{tabular}{
    l      
    c      
    c      
    c      
    c      
    c      
    c      
    c      
    c      
    c      
    c      
}
\toprule[1.3pt]
\textbf{Dataset} & \textbf{Year} & \textbf{Resolution} & \textbf{Modality} &
\textbf{Boxes}  & \textbf{Classes} & \textbf{Manual} & \textbf{Real} & \textbf{HS} & \textbf{LL} ($\mu Y_{709}<1$)  &  \textbf{Ext.(\%)} \\
\midrule[1pt]
    Gen1 & 2020 & $304 \times240$ & Event & 255K & 2 & \cmark & \cmark & \xmark & \cmark & - \\
    1 Mpx & 2020  & $1280\times720$ & Event & 25M & 7 & \xmark & \cmark & \xmark & \cmark  & - \\
   
    PEDRo & 2023 & $346 \times260$ & Event & 43K & 1 & \cmark & \cmark & \xmark & \xmark & - \\
    eTraM & 2024 & $1280\times720$ & Event & 2M & 3 & \cmark & \cmark & \xmark & \cmark  & -  \\
    SEVD & 2024 & $800 \times600$ & Event & 9M & 6 & \xmark & \xmark & - & -  & - \\
    Event-KITTI & 2024 & $1333\times401$ & Event &  80K & 8 & \cmark & \xmark & - & - & -\\
    DSEC & 2021 & $640 \times480$ & Frame, Event & 390K & 8 & \xmark & \cmark  & \xmark & \xmark  & 20 \\
    PKU-SOD & 2022 & $346 \times260$ & Frame, Event & 1080K & 3 & \xmark & \cmark & \cmark & \xmark  & 13 \\
   
\midrule[1.3pt]
    \textbf{PEOD(Ours)} & 2025 & $\mathbf{1280}\times\mathbf{720}$
 & \textbf{Frame, Event} & \textbf{340K} & 6 & \cmark & \cmark & \cmark & \cmark & \textbf{57} \\
 
\bottomrule[1.3pt]
\end{tabular}
\caption{\textbf{Comparison with existing object detection datasets.} PEOD dataset is the first to provide a high-resolution, 6-class benchmark annotated at 30Hz, with over 57\% of data focusing on extreme scenarios. 
   LL: Mean BT.709 luminance of frames ($\mu Y_{709}$) in the 0-255 domain.
    HS: High-speed scenarios.
    Ext.(\%): Proportion of sequences collected under extreme conditions. }
\label{tab:dataset_compare}
\end{table*}
}

\subsection{Data Statistics}
The PEOD dataset consists of over 130 driving sequences, up to 90s in duration. It contains over 72k annotated frames, totaling 340k bounding box labels across the six categories. For our experiments, we partition the dataset into a training set of 270k boxes and a test set of 70k boxes.

Motion‑blurred and sharp frames frequently coexist within the same driving sequence, and our statistics indicate that roughly 40-50\% of frames in nominally high‑speed segments are blur‑free. Assigning blur labels at the frame level would fragment temporal context and compromise sequence‑wise evaluation. Therefore, we center our dataset split on illumination, the factor where RGB sensors degrade the most yet event cameras excel. To quantitatively identify and categorize these conditions, we define an underexposure score $(S_{LL})$ and an overexposure score $(S_{OE})$ based on the pixel saturation in each grayscale frame $F$. The formulations are as follows:

\begin{equation}
S_{LL} = \frac{1}{W \times H} \sum_{i=1}^{W} \sum_{j=1}^{H} \mathbf{I}(F_{ij} < T_{\text{dark}})
\label{eq:SLL}
\end{equation}

\begin{equation}
S_{OE} = \frac{1}{W \times H} \sum_{i=1}^{W} \sum_{j=1}^{H} \mathbf{I}(F_{ij} > T_{\text{bright}})
\label{eq:SOE}
\end{equation}

\noindent where \textit{I} is the indicator function, and $T_{dark}$ and $T_{bright}$ are predefined thresholds for dark and bright saturation, respectively. Using thresholds of 30 and 250, a frame is classified into a specific subset (e.g., low-light, overexposed) if its corresponding score exceeds a certain percentage threshold. 
Based on illumination conditions, we divide the dataset into two subsets: \textbf{1) \textit{Illumination Challenge Subset}}: sequences recorded under challenging illumination conditions (e.g., low-light, overexposure, abrupt lighting changes ). \textbf{2) \textit{Normal Subset}}: sequences recorded under standard lighting conditions. The approximate distribution across these subsets is 57.1\% extreme lighting scenarios and 42.9\% normal lighting scenarios.

\subsection{Comparison with other Datasets}

{
\setlength{\tabcolsep}{4pt}
\begin{table*}[t]
\centering
\begin{tabular}{
    l      
    c      
    c      
    c      
    c      
    c      
    c      
    c      
    c      
    c      
}
\toprule[1.3pt]
\textbf{Input} & \textbf{Method} & \textbf{Pub.\ \& Year} & \textbf{Backbone} &
\textbf{mAP} & \textbf{mAP$_{50}$} & \textbf{mAP$_{75}$} &
\textbf{Param(M)} & \textbf{FLOPs(G)} & \textbf{T(ms)} \\
\midrule
\midrule
\multirow{6}{*}{Event}
    & YOLOX(Event)            & arXiv'21 &  CSPDN & 16.1  & 28.3 & 16.0 & 8.9 & 32.1 & 6.5 \\
  & RVT              & CVPR'23 & Transformer & 20.1  & 38.4 & 18.8 & 4.4 & 31.2 & 9.3 \\
  & SAST             & CVPR'24 & Transformer & 18.1  & 37.8 & 16.7 & 4.5 & 37.1 & 19.1 \\
   & SpikingYOLO     & ECCV'24 & SNN    & 10.2 & 21.8 & 7.9    & 23.1 & 136.7 & 53.8 \\
  & SMamba           & AAAI'25 & SSM & 22.9     & 43.8    & 19.9    & 23.7 & 72.8 & 38.7 \\

\midrule
\multirow{4}{*}{RGB}
& RetinaNet           & arXiv'20 &  ResNet-50                  & 12.6     & 22.1  & 12.5  & 36.4  & 198.2 & 14.5 \\
  & YOLOX            & arXiv'21 &  CSPDN                    & 17.4  & 34.0 & 14.2 & 8.9 & 32.1 & 6.5 \\
  & YOLOv7           & CVPR'23 &  CSPDN                 & 19.8     & 37.1    & 19.1   & 6.1  & 31.5 & 6.17 \\
  & YOLOv8           & 2023 &  CSPDN                 & 18.9     & 36.3    & 18.8    & 11.1 & 34.4 & 5.4 \\
  & RT-DETR           & CVPR'24 &  ResNet-18                 & 21.8     & 38.5    & 21.3    & 23.1 & 61.6 & 10.9 \\
\midrule
\multirow{6}{*}{Event+RGB}
  & FPN-Fusion       & ICRA'22 & ResNet-50 & 14.6 & 29.9 & 12.6 & 65.6 & 283.7 & 32.2 \\
  & RENet            & ICRA'23 & CSPDN & 24.5 & 41.6 & 22.8 & 37.7 & 102.7 & 28.5 \\
  & SODFormer        & TPAMI'23 & ResNet-50      & 17.8 & 36.3 & 15.0 & 86.5 & 279.7 & 48.7 \\
  & EOLO             & ICRA'24 & SNN + CSPDN & 26.1 & 45.8 & 26.2 & 46.2 & 100.9 & 22.6 \\
  & SFNet            & TITS'24 &  CSPDN & 19.5 & 37.9 & 17.7 & 38.7 & 103.8 & 23.8 \\
  & CAFR              & ECCV'24  & ResNet-50 & 21.4 & 39.6 & 19.3 & 82.1 & 319.9 & 52.4 \\
\bottomrule[1.3pt]
\end{tabular}
\caption{Results on the PEOD benchmark comparing RGB, Event, and Event-RGB detectors. (CSPDN denotes the CSPDarknet backbone). The fusion detectors show a clear performance advantage, yielding the highest mAP over single-modality methods.}
\label{tab:benchmark}
\end{table*}
}

In Table \ref{tab:dataset_compare}, we compare our PEOD dataset with other object detection datasets. In contrast, other large-scale public event-based datasets, such as Gen1 \cite{Gen12020} and 1 Mpx dataset \cite{1MPX2020}, offer only long-duration event streams. This limitation hinders the development of high-precision, all-day object detection systems, particularly in static or extremely slow-moving scenarios. Datasets like PEDRo \cite{Boretti2023} and eTram \cite{Verma2024} are tailored for niche applications, focusing on event-based pedestrian detection and traffic flow monitoring from a surveillance perspective, respectively. Furthermore, while SEVD \cite{SEVD2024} and Event-KITTI \cite{EVENTKTTI2024} provide large-scale event data, they rely on event simulators, creating a significant domain gap compared to data captured by real-world event cameras. More importantly, all the aforementioned datasets exclusively provide event streams. Among datasets that offer aligned RGB and event data, DSEC \cite{dsecGehrig2021} and PKU-SOD \cite{Li2023} are constrained by their low spatial resolution, which limits the performance of detection algorithms that require fine-grained detail. Moreover, challenging scenarios constitute less than 20\% of their data, rendering them inadequate for a comprehensive evaluation of model robustness under adverse conditions.

Overall, our PEOD dataset offers four key advantages:
\textbf{1) Challenging Scenarios:} Over 57\% of the dataset comprises sequences captured in difficult conditions.
\textbf{2) High Spatiotemporal Resolution:} The dataset features high spatial resolution of 1280$\times$720, complemented by the microsecond-level temporal resolution.
\textbf{3) High Dynamic Range:} The event camera ensures an HDR of over 120dB, preserving signal integrity in extreme Illumination conditions.
\textbf{4) Dense and Diverse Annotations:} The data stream is continuously annotated with 6 object classes at 30Hz. 

\section{Evalution and Benchmark}
\subsection{Experimental Setup}
All experiments are conducted on our proposed PEOD dataset, adhering strictly to the official train-test splits detailed in Section 3. In addition to evaluating model performance on the complete test set, we specifically assess robustness under distinct illumination conditions using the Illumination Challenge and Normal subsets. Three distinct categories of detectors are systematically evaluated: Event-based, RGB-based, and Event-RGB fusion approaches. For Event-based models such as RVT \cite{Gehrig2023}, SAST \cite{Peng2024}, and SMamba \cite{Yang2025}, asynchronous event streams are encoded into tensor representations using a stacked-histogram method \cite{Gehrig2023}, accumulating events at the original spatial resolution within a fixed 33 ms temporal window partitioned into 10 bins. These models are uniformly trained for 120k iterations. For the SNN-based detector (SpikingYOLO), We strictly follow its original methodology by collecting events within the 250ms preceding each annotation, evenly splitting this interval into two temporal segments for event integration. For RGB-based and Event-RGB fusion detectors, events are first transformed into event images with an integration window of 33ms. Both RGB images and event images are uniformly resized to a resolution of $768\times1280$ before passing through their respective feature extractors. 
Detectors in these two categories are trained for 20 epochs. All three categories (Event-based, RGB-based, and Event-RGB fusion) are trained on 4$\times$ NVIDIA RTX 4090 GPUs. Hyperparameter  and training settings mostly follow the original configurations.



\textbf{Evaluation Metrics.} To enable fair comparison across three-class detectors, we report 1) mean Average Precision (mAP) on the COCO benchmark, 2) inference time per image (ms), and 3) model size measured by the number of parameters. For mAP, we provide scores at multiple IoU thresholds: $\text{mAP}_{50{:}95}$, $\text{mAP}_{50}$, and $\text{mAP}_{75}$.

\subsection{Benchmark Evalution }

We conduct a comprehensive evaluation on our PEOD dataset, comparing three categories of detectors. The results, summarised in Table \ref{tab:benchmark}, quantitatively characterize the performance of each category.

\subsubsection{Evaluation on Event-based Detectors.} To evaluate performance on the event stream, we benchmark several detectors that represent distinct model classes, including Transformer-based (RVT and SAST), CNN-based (YOLOX) \cite{yolox2021}, SNN-based (SpikingYOLO) \cite{Luo2024}, and Mamba-based (SMamba). As reported in Table \ref{tab:benchmark}, SMamba attains the highest detection accuracy of 22.9\%, whereas RVT and SAST exhibit marginally lower performance. Although SMamba clearly excels at capturing long-term temporal dependencies, its 38.7ms inference latency and 72.8 GFLOPs indicate that the enhanced representational capacity is achieved at the cost of considerably greater computational overhead. By contrast, SpikingYOLO achieves limited detection accuracy, a limitation attributable to the still-maturing training frameworks for SNN and their specialized hardware requirements.

\subsubsection{Evaluation on RGB-based Detectors.} We benchmark several classical frame-based object detectors, including RetinaNet \cite{Lin2017Retinanet} RT-DETR \cite{DETR2024} and multiple YOLO variants \cite{yolov72023,yolov82023,yolox2021}, which frequently serve as baselines for subsequent fusion strategies. RT-DETR achieves the best performance among these models, albeit with higher latency (10.9ms) and more parameters (23.1M).  On the other hand, the YOLO variants present a more balanced performance profile. The results obtained by these frame-based detectors underscore the effectiveness of conventional cameras under favorable conditions where rich texture and color information are available.

\subsubsection{Evaluation on Event-RGB Fusion Detectors.} We evaluate 6 fusion detectors, FPN-Fusion \cite{Tomy2022}, RENet \cite{Zhou2022}, SODFormer \cite{Li2023}, EOLO \cite{Cao2024EOLO}, SFNet \cite{Liu2024} and CAFR \cite{Cao2024}. Integrating event data with RGB frames produces substantial performance gains relative to single-modality detectors. EOLO achieves the highest detection accuracy of 26.1\%, representing an absolute improvement of 4.3\% over RGB-based detectors and 3.2\% over event-based model. Comparative analysis shows that architectures such as EOLO and RENet, which employ modality-aware fusion mechanisms, markedly surpass simple feature concatenation detectors exemplified by FPN-style fusion. These results underscore that sophisticated schemes capable of jointly leveraging frame texture features and motion cues captured by event streams are indispensable for realizing the full potential of multimodal input.


Across the three detector categories, fusion models exhibit higher latency than RGB-based detectors. However, techniques such as EOLO (22.6ms) and SFNet (23.8ms) still provide practical inference speeds. Consequently, Event-RGB fusion constitutes an effective means of attaining significant improvements in detection accuracy.

\subsection{Condition-Specific Evaluation}
A comparative performance analysis of three detector categories was conducted on the Illumination Challenge and Normal subsets. The comprehensive results are reported in Table \ref{tab:illum} and Table \ref{tab:normal}, respectively.

{
\setlength{\tabcolsep}{4pt}
\begin{table}[htbp]
\centering
\begin{tabular}{l l c c c }
\toprule[1.3pt]
\textbf{Input} & \textbf{Method} & \textbf{mAP} & \textbf{mAP$_{50}$} & \textbf{mAP$_{75}$}  \\
\midrule
\midrule
\multirow{6}{*}{Event}
& YOLOX(Event)        & 17.5 & 33.1 & 16.3  \\
  & RVT          & 20.4 & 39.1 & 19.0  \\
  & SAST         & 19.1 & 38.5 & 16.5  \\
  & SpikingYOLO & 10.2 & 22.6 & 7.2  \\
  & SMamba & 23.2 & 44.5 & 20.3  \\
\midrule
\multirow{4}{*}{RGB}
& RetinaNet      & 8.0  & 16.4 & 7.0 \\
  & YOLOX        & 10.5 & 24.6 & 8.0  \\
  & YOLOv7 & 12.6 & 31.1 & 13.0  \\
  & YOLOv8 & 11.9 & 27.7 & 11.1  \\
  & RT-DETR & 13.2 & 30.1 & 13.9  \\
\midrule
\multirow{6}{*}{Event+RGB}
  
  & FPN-Fusion & 11.2 & 23.6 & 9.0  \\
  & RENet & 10.8 & 24.4 & 8.3  \\
  & SODFormer & 10.4 & 23.6 & 7.7 \\
  & EOLO & 11.5 & 25.2 & 9.2  \\
  & SFNet        & 11.3 & 28.1 & 8.6 \\
  & CAFR & 11.2 & 25.2 & 8.6  \\
\bottomrule[1.3pt]
\end{tabular}
\caption{Results on Illumination Challenge Subset (e.g., low-light, overexposure, abrupt illumination changes) comparing RGB-based, Event-based, and Event-RGB detectors.}
\label{tab:illum}
\end{table}
}

Experimental results demonstrate the superior robustness of event-based detectors. Under both normal and extreme illumination, the event-based detectors maintain essentially constant performance, attributable to the high dynamic range of event cameras and their asynchronous, sparsity-driven data representation. By contrast, RGB-based detectors suffer a pronounced drop when exposed to low-light, overexposed, or high-speed conditions, where conventional frame sensors yield degraded images. In the challenging illumination scenarios, this image degradation explains the sharp decline in RGB-based detectors. 

However, Event-RGB fusion detectors outperform their corresponding RGB baseline by approximately 2\% in mAP, effectively compensating for missing appearance cues with event-derived edge information. Conversely, in normal scenes, rich color and texture allow RGB-based detectors to reach high accuracy, while the event-based detectors underperform because of their limited fine-grained appearance cues. Event-RGB fusion detectors attain the top performance by integrating rich texture information in RGB frames with the motion cues inherent in event data.

{
\setlength{\tabcolsep}{4pt}
\begin{table}[htbp]
\centering
\begin{tabular}{l l c c c }
\toprule[1.3pt]
\textbf{Input} & \textbf{Method} & \textbf{mAP} & \textbf{mAP$_{50}$} & \textbf{mAP$_{75}$}   \\
\midrule
\midrule
\multirow{5}{*}{Event}
& YOLOX(Event)    & 14.0 & 23.2 & 12.0 \\
  & RVT          & 19.0 & 36.0 & 18.1 \\
  & SAST         & 16.7 & 35.6 & 17.4  \\
  & SpikingYOLO & 10.1 & 20.1 & 9.7 \\
  & SMamba & 21.9 & 41.2 & 19.1 \\
\midrule
\multirow{4}{*}{RGB}
& RetinaNet      & 20.0 & 32.5 & 21.7 \\
  & YOLOX        & 31.7 & 49.8 & 28.7 \\
  & YOLOv7         & 32.9 & 50.5 & 32.5 \\
  & YOLOv8 & 29.3 & 49.2 & 30.6 \\
  & RT-DETR & 36.5 & 51.2 & 37.6 \\
\midrule
\multirow{6}{*}{Event+RGB}
  & FPN-Fusion & 21.3 & 37.8 & 21.8  \\
  & RENet & 43.7 & 65.8 & 43.7   \\
& SODFormer & 30.8 & 54.0 & 28.8\\
  & EOLO & 45.2 & 66.7 & 48.4  \\
  & SFNet        & 38.9 & 58.3 & 41.3\\
  & CAFR & 28.5 & 47.9 & 29.9 \\
\bottomrule[1.3pt]
\end{tabular}
\caption{Results on the Normal Subset comparing RGB-based, Event-based, and Event-RGB detectors. }
\label{tab:normal}
\end{table}
}

In normal illumination, fusion detectors effectively integrate the complementary strengths of RGB and event streams, yielding a substantial improvement in detection accuracy. In extreme illumination, fusion detectors still exceed RGB-based detectors but remain inferior to the strongest event-based detectors. The divergent performance across the two subsets exposes intrinsic weaknesses in current fusion strategies: \textbf{1)} Most architectures rely on shallow feature concatenation, lacking mechanisms to suppress the noise introduced by degraded RGB. \textbf{2)} During training, the weighting scheme disproportionately favors the texture-rich RGB branch, so event features cannot take precedence when lighting conditions deteriorate. \textbf{3)} Existing designs fail to exploit the high temporal resolution and motion cues that are unique to event data. Closing the performance gap in adverse scenarios will require reliability-aware, deeply coupled frameworks that adaptively reweight the two modalities and fully leverage event representations, thereby delivering consistent and robust perception under all illumination conditions.

\subsection{Qualitative Analysis}

\begin{figure*}[t]
    \centering
    \includegraphics[width=1\textwidth]{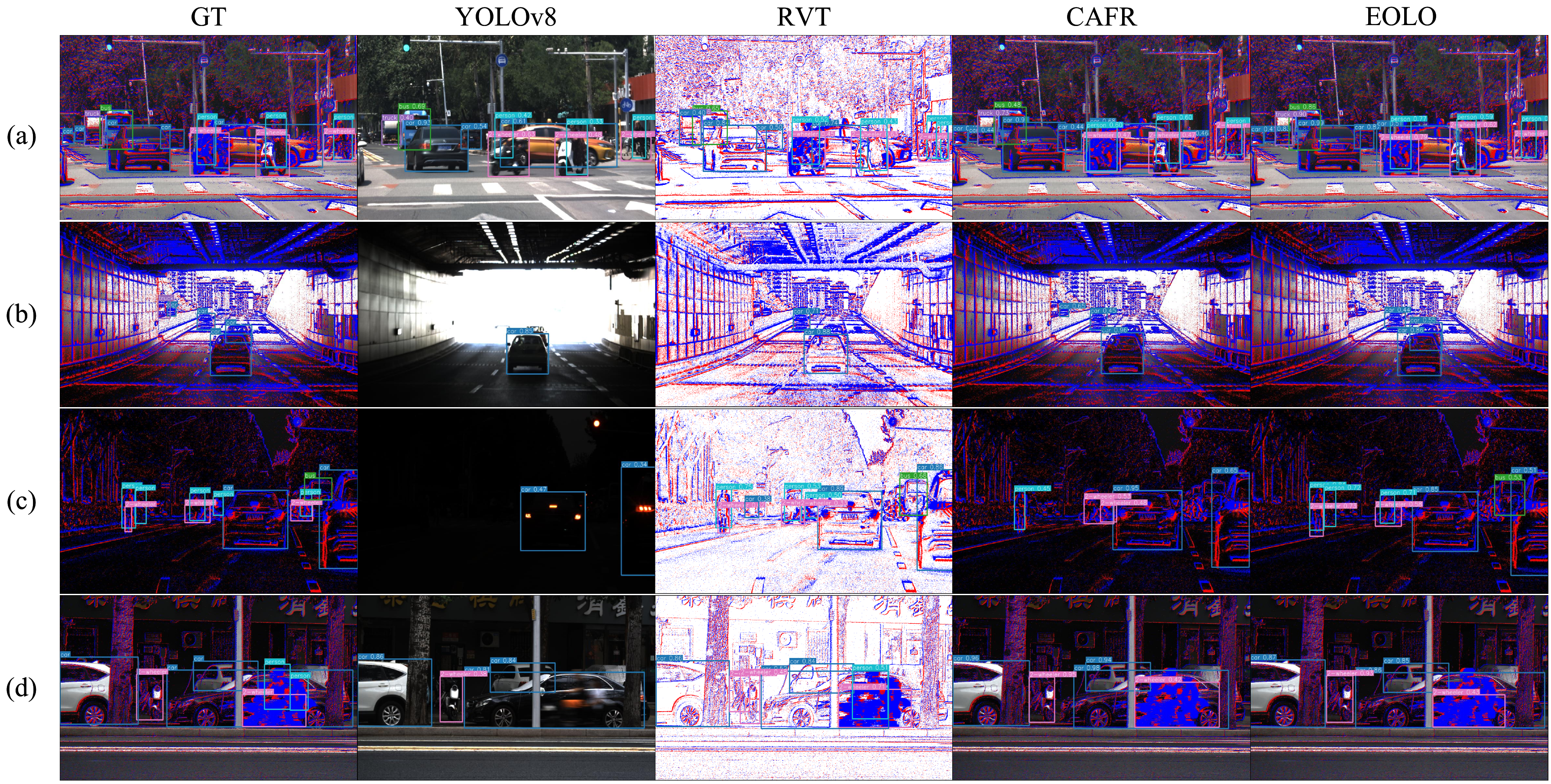}
    \caption{Representative visualization results on our PEOD dataset. (a) Traffic intersection in normal scenario. (b) Rushing cars in overexposure scenario. (c) Traffic intersection in low-light scenario.
    (d) High-speed moving two-wheelers with motion blur. While the RGB-based detector (YOLOv8) effectively utilizes rich textures in the normal scene (a), fusion detectors (CAFR, EOLO) and the event-based detector show decisive advantages by leveraging event data in the challenging overexposure (b), low-light (c), and motion-blur (d) conditions where the RGB-based detector fails.
    }

    \label{fig:ex_show}
\end{figure*}

Figure \ref{fig:ex_show} contrasts representative results from an RGB-based detector (YOLOv8), an event-based detector (RVT), and two Event-RGB fusion detectors (CAFR and EOLO), alongside the ground truth. The analysis spans four characteristic scenes: (a) a normal traffic intersection; (b) a tunnel exit exhibiting severe overexposure; (c) a low-light intersection; and (d) high-speed two-wheelers affected by motion blur. These qualitative results substantiate the scenario-specific conclusions discussed earlier. \textbf{(a) Normal scene.} In well-illuminated conditions, RGB frames provide rich texture and color cues, enabling the RGB-based detector to localize most medium-sized objects reliably. In contrast, the event stream contains sparse edge-like responses and limited texture, which leads the event-based model to miss small or heavily occluded instances and to produce fragmented boxes. Fusion detectors leverage complementary cues and recover several small or partially occluded targets that are absent from the event-based predictions, while maintaining precise localization. \textbf{(b) Overexposure.} At the tunnel exit, saturation in RGB frames suppresses object contrast and the RGB-based detector consequently fails to detect multiple vehicles. Event data, however, remains informative due to its high dynamic range, allowing the event-based detector to retain object contours and maintain detections. Fusion detectors inherit this robustness: both CAFR and EOLO correctly detect vehicles that the RGB-based detector misses, illustrating how event cues compensate for severe illumination degradation. \textbf{(c) Low-light.} Under nighttime conditions, RGB frames lose contrast and texture, causing the RGB-based detector to miss targets. The event-based detector benefits from strong responses to intensity changes (e.g., headlights, motion edges) and therefore detects the main traffic participants. Fusion produces competitive results, but when the RGB modality is extremely degraded, rigid fusion can inject noise from the RGB branch and slightly undercut the discriminative signal of the event stream. This modality-conflict effect is consistent with our quantitative results on the Illumination Challenge subset, where current fusion strategies may underperform strong event-based baselines. \textbf{(d) Motion blur.} Pronounced blur in RGB frames smears spatial details and leads to missed or poorly localized two-wheeler instances for the RGB-based detector. Event streams preserve sharp motion-induced edges at microsecond resolution, enabling the event-based detector to retain detections. Fusion further improves the stability and tightness of bounding boxes by synergizing motion cues from the event stream with the residual textural or structural information still discernible in the blurred RGB frame.

These qualitative examples highlight the complementary nature of events and frames: RGB excels in texture-rich, well-lit scenes but degrades under saturation, darkness, or blur, while events remain reliable in those adverse conditions yet struggle with small, distant, or slow targets due to weak texture and sparse contours. By furnishing pixel-aligned Event-RGB pairs captured across a spectrum of extreme lighting and motion scenarios, PEOD supplies a large-scale, high-fidelity dataset and benchmark for systematically closing the Event-RGB fusion gap 
under such conditions.


\subsection{Discussion and Outlook}

Rich in challenging scenarios, PEOD promises to catalyze progress across key tasks: \textbf{(1) Image Reconstruction.} Learning-based event-to-image reconstruction is hampered by synthetic training data that fail to generalize to real scenes, while PEOD bridges this sim-to-real gap by providing a dataset of pixel-aligned Event-RGB pairs.
\textbf{(2) Object Tracking.} PEOD provides long, continuous real-world sequenceswith occlusions and extreme lighting, enabling evaluation of long-term identity preservation, robustness, and real-time efficiency.

\section{Conclusion}
We introduce a high-resolution, pixel-aligned Event-RGB dataset PEOD and benchmark for object detection, with extensive coverage of extreme scenarios. By addressing the inadequate resolution and scarcity of adverse conditions found in existing datasets, PEOD provides a solid foundation for future robust perception. Our comprehensive evaluation indicates that while fusion detectors achieve the best overall performance, they still fail to fully leverage the event stream when frame is severely degraded, highlighting the need for more sophisticated fusion strategies. Furthermore, PEOD enables image reconstruction for bridging the sim-to-real gap, while its challenging sequences advance robust multi-object tracking especially extreme illumination shifts.


\section{Acknowledgments}
This work was supported by BUPT Tnnovation and Entrepreneurship Support Program (2025-YC-T026); the National Key R\&D Program of China (2021ZD0109802); High-performance Computing Platform of BUPT.
\bibliography{aaai}

\end{document}